# Residual Unfairness in Fair Machine Learning from Prejudiced Data

Nathan Kallus [1]  Angela Zhou [2]

## Abstract

Recent work in fairness in machine learning has proposed adjusting for fairness by equalizing accuracy metrics across groups and has also studied how datasets affected by historical prejudices may lead to unfair decision policies. We connect these lines of work and study the *residual unfairness* that arises when a fairness-adjusted predictor is not actually fair on the target population due to systematic censoring of training data by existing biased policies. This scenario is particularly common in the same applications where fairness is a concern. We characterize theoretically the impact of such censoring on standard fairness metrics for binary classifiers and provide criteria for when residual unfairness may or may not appear. We prove that, under certain conditions, fairness-adjusted classifiers will in fact induce residual unfairness that perpetuates the same injustices, against the same groups, that biased the data to begin with, thus showing that even state-of-the-art fair machine learning can have a "bias in, bias out" property. When certain benchmark data is available, we show how sample reweighting can estimate and adjust fairness metrics while accounting for censoring. We use this to study the case of Stop, Question, and Frisk (SQF) and demonstrate that attempting to adjust for fairness perpetuates the same injustices that the policy is infamous for.

## 1. Introduction

The spread of data-driven decision making to civic institutions, spurred by the empirical success of machine learning and the growing availability of individual-level data, raises new questions about the possible harms of learning from data which is subject to historical bias. Unlike clean-cut prediction problems in other domains, datasets of individuals and their historical outcomes may reflect systematic bias due to previously prejudiced decisions (Barocas & Selbst, 2014). Recent work on fairness in machine learning proposes and analyzes competing criteria for assessing the fairness of machine learning algorithms, where some adjustments attempt to equalize accuracy metrics across groups (Corbett-Davies et al., 2017; Kleinberg et al., 2017; Hardt et al., 2016). Other work studies how historical prejudices may be reflected in training data such that algorithmic systems might replicate historical biases (Angwin et al., 2016; Lum & Isaac, 2016; Kilbertus et al., 2017). We consider a model of biased data where systematic censoring affects whether or not entire observations appear in the training dataset. In such cases, the available data are not representative of the eventual real-world "test" population to which any resulting learned policy will be applied. Our paper formalizes and characterizes how systematic under- or over-representation of groups in the dataset can hamper attempts to correct for fairness, leading to residual unfairness on the target population of interest.

This important issue arises in almost all settings where fair machine learning has been studied:

(1) Data on loan default can only be collected on those loan applicants who were historically approved but is used to learn approval policies applied to all applicants.
(2) Arrest data help build predictive policing models but these data are disproportionately collected on individuals in highly patrolled areas and may be subject to further prejudice at the individual level, including racial (Lum & Isaac, 2016).
(3) Risk assessment and intervention tools in child welfare agencies are trained on cases which have been "screened in" by caseworkers based on external reports (Chouldechova et al., 2018).
(4) Defendants may only flee and fail to appear if not detained, so any flight risk score used for setting bail can only be learned from data on defendants who were not detained.
(5) Convict recidivism is impacted by sentence applied but learned risk scores are applied to all convicts.

In these applications, systematic censoring screens out ob-

This material is based upon work supported by the National Science Foundation under Grant No. 1656996. Angela Zhou is supported through the National Defense Science & Engineering Graduate Fellowship Program. [1]Cornell University, and Cornell Tech, New York [2]Cornell University, Ithaca, New York. Correspondence to: Angela Zhou <az434@cornell.edu>.

*Proceedings of the 35$^{th}$ International Conference on Machine Learning*, Stockholm, Sweden, PMLR 80, 2018. Copyright 2018 by the author(s).

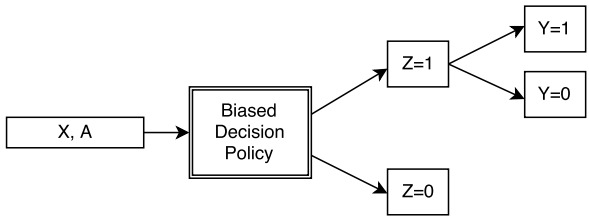

Figure 1: Problem Setting: Censored Data. Outcomes are only observed for those included in the dataset ($Z = 1$).

servations of individuals and their outcomes from a training dataset. Such censoring may reflect historical decisions made with limited access to information, heterogeneous decision-makers, or the application of statistically discriminatory rules (Arrow, 1973). Despite the intermediate screening, domain-level restrictions may require ensuring fairness of any decision or prediction policy with respect to the original population. We formalize this mechanism as a data setting (Fig. 1) where a historical decision policy $Z \in \{0, 1\}$ specifies whether an instance will be included in the dataset. Systematic censoring may induce *covariate shift* on population-level estimands, such as true positive rate, as outcomes are observed in the training data only where $Z = 1$. Notably, predictive tools built on these censored datasets are actively being deployed: there is an opportunity to improve upon standards of practice, but the structural implications of systematically censored data ought to be accounted for (Angwin et al., 2016; Capatosto, 2017; LJAF, 2015). Our contributions are as follows:

- We characterize when systematic censoring induces *residual unfairness* in terms of the distributions of the conditional Bayes-optimal risk score across censored and target groups.
- When benchmark data is available, we show how to use sample re-weighting techniques to estimate accuracy metrics to adjust for fairness on the target population. We show how the sample weights indicate what groups remain disadvantaged by residual unfairness.
- We demonstrate how systematic inclusion can affect fairness adjustments on an empirical example with data from the application of the Stop, Question, and Frisk (SQF) policy of the City of New York Police Department (NYPD).

In settings where datasets can be subject to historical prejudice and decision policies ought to be truly fair on the general population, we argue it is paramount to carefully consider and account for the sampling process to ensure fairness on the true population.

## 2. Problem Setting

We consider the problem of learning a fair decision policy (classifier or threshold rule on a regressor) from a dataset where each decision instance is characterized by observed covariates $X \in \mathcal{X}$ (e.g., predictors of creditworthiness, criminality, etc), protected class $A \in [m] = \{1, \ldots, m\}$ (e.g., sex, race, etc), and label $Y \in \{0, 1\}$ (where $Y = 1$ is generally interpreted as the favorable label, e.g., "will pay back loan" or "will not reoffend").

Fig. 1 illustrates the construction of a biased training dataset in this setting. The indicator $Z \in \{0, 1\}$ specifies whether an instance is included in the post-censoring training data (e.g., "approved for a loan") and another indicator $T \in \{0, 1\}$ specifying whether an instance belongs to the population to which the learned policy will be applied. For example, if $T$ is the constant 1, the target population is that of all loan applicants. We sometimes call $Z = 1$ the *logging policy* in analogy to logged-bandit learning (Kallus, 2017; Swaminathan & Joachims, 2015), where the implementation of a (often unknown) historical policy resulted in limited bandit feedback on outcomes. Because a random sample from the target population is generally not available, the target population is different from the notion of a held-out "test" dataset used to evaluate predictive accuracy.

We consider the problem of determining a policy assigning labels $\hat{Y} \in \{0, 1\}$ that depends only on $X, A$ but may be randomized (so $\hat{Y} \perp\!\!\!\perp (Y, Z, T) \mid X, A$). Labeled training data $(X, A, Y)$ is available from the $Z = 1$ population, so that only the conditional joint distribution of $X, A, Y \mid Z = 1$ is characterized by this data. We may or may not also have unlabeled data from the $T = 1$ population, $X, A \mid T = 1$.

For concreteness, when discussing fairness criteria, we consider the specific fairness criterion of equality of opportunity or equalized odds introduced by Hardt et al. (2016). The adjustment determines a fair policy $\hat{Y}$ from a (possibly discriminatory) black-box binary predictor or score $\hat{R}$ without access to the original training data. They identify two particular types of fairness, equal opportunity and equalized odds, which require that a fairness-adjusted policy $\hat{Y}$ be independent of class $A$ given a positive label $Y = 1$ or given any label $Y$, respectively. For loan approval, equality of opportunity requires that the policy treat truly creditworthy individuals the same, independent of protected class membership. Equalized odds prohibits abusing class membership as an unfair proxy for $Y$ (e.g., via stereotyping or racial profiling). For our setting, to be explicit, we define these relative to an event:

**Definition 1.** A policy $\hat{Y}$ satisfies *equalized odds* with respect to (wrt) class variable $A$ and event $E$ if

$$\hat{Y} \perp\!\!\!\perp A \mid Y = y, E \qquad (1)$$

holds for $y \in \{0, 1\}$. A policy satisfies *equal opportunity* wrt $A$ and $E$ if eq. (1) holds for $y = 1$.

For brevity, we will say a policy is simply *fair* to mean either equal opportunity or equalized odds. Hardt et al.

(2016) determine such a policy by a post-processing step given a score $\hat{R}$ using a constrained optimization problem over group-specific thresholds (potentially randomized), enforcing the constraints in eq. (1) on the true positive rate and/or false positive rate across groups while optimizing a given classification loss. Specifically, define $F_a^E(\theta) = \Pr[\hat{R} \leq \theta \mid Y = 1, A = a, E]$ as the conditional CDF (cumulative distribution function) of the score given the event $E$ and $Y = 1$. For a given true positive rate $\rho$, the corresponding *derived equal opportunity classifier at rate* $\rho$ is given by $\hat{Y} = \mathbb{I}[\hat{R} > \theta_A]$, where $\theta_a = (F_a)^{-1}(1 - \rho)$ is the threshold corresponding to group $a$ so that it has true positive rate $\rho$. Note that $F_a^{Z=1}, F_a^{T=1}$ are the false negative rates for a threshold classifier as evaluated on the censored and target population, respectively. Naturally, a policy is actually fair if it is fair on the population to which it is applied (here, $T = 1$). So, in seeking a fair policy per these definitions, we seek a policy that satisfies equal opportunity or equalized odds on the target population wrt class variable $A$ and the event $T = 1$, while the approach of Hardt et al. (2016) applied directly to training data achieves fairness wrt $A$ and the event $Z = 1$. Throughout, we assume $\hat{R} \in [0, 1]$.

## 3. Related Work

**Fairness and Missing Data.** Research on fairness and machine learning has considered some subcomponents of the overall problem we study of learning fair policies from biased datasets. Hardt et al. (2016) formalize the criteria of equal opportunity and equalized odds. Lum & Isaac (2016) show that a predictive policing algorithm for drug enforcement in Oakland, trained on police records, will perpetuate disparate enforcement. Ensign et al. (2017) consider a discrete model of how beliefs of crime rates in different areas adjust after observing arrest rates, and propose implementing the Horvitz-Thompson estimator via rejection sampling of arrest observations in an "online" fashion. Lakkaraju et al. (2017) identify a similar structural setting with "selective labels" where they learn a decision rule for pre-trial risk assessment from the decisions made from judges (which affect whether or not the outcome of interest will be observed). They leverage the heterogeneity of heterogeneous decision makers using different decision thresholds to identify subpopulations for comparison but do not consider the subsequent fairness of the learned policy.

**Sampling Adjustment and Superpopulations.** Sampling adjustment and re-weighting is commonly used in the social sciences, medicine, and epidemiology for ensuring the validity of population-level inference where there is population mismatch between studies and the population of interest (Thompson, 2012; Freedman & Berk, 2008). The classic Horvitz-Thompson estimator uses the inverse probability of sampling probability weights and is unbiased for population level estimates (Horvitz & Thompson, 1952). Much of the work on fairness in machine learning has used population-level statistics such as accuracy metrics (true positive rate, false negative rate) as metrics for identifying disparate impact. The case of sample selection bias was studied in Zadrozny (2004) for classifier evaluation, without regard for fairness impacts.

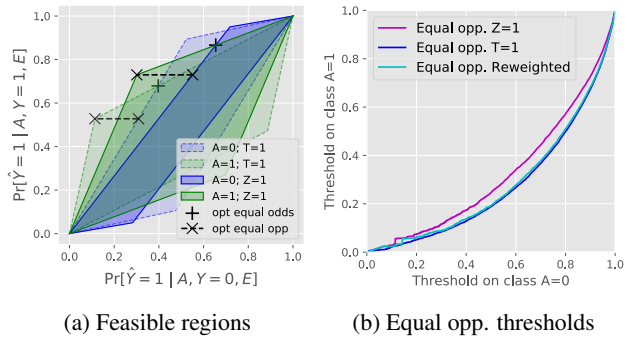

(a) Feasible regions  (b) Equal opp. thresholds

Figure 2: Illustrative synthetic example: Comparison of equal-opportunity or equal-odds adjustments derived from censored or target population data.

## 4. Residual Unfairness Under Disparate Benefit of the Doubt: Bias In, Bias Out

We study how prejudicial biases in a dataset can lead to *residual unfairness*, which persists even after fairness adjustment if error parity metrics assessed from the censored dataset are used. We show that the residual unfairness that remains even after adjustment will disadvantage the same group that was prejudiced against before, in the training data. This proves that even after fairness adjustment, fair machine learning still has a "bias in, bias out" property.

**An Illustrative Synthetic Example: Loan Application** To illustrate the potential impact of population mismatch on fairness adjustments in a controlled setting, we consider a synthetic example for loan approval. Suppose there are two classes, with half of the population of loan applicants in class 0 and the other half in class 1. We let $T = 1$ be constant as we wish to consider the impact of our policy on the whole population of loan applicants. We denote by $X_1$ the (normalized) number of bank accounts and by $X_2$ the (normalized) number of delinquent payments on record, including those for subprime loans.[1]

Suppose $X$ is distributed as a standard bivariate normal conditioned on class, with a mean of $(1, 0)$ among individuals in the class $A = 0$ and a mean of $(0, 1)$ among individuals in the underprivileged class $A = 1$. Consider $Y \in \{0, 1\}$ indicating whether the individual will pay back the loan if it were approved. Suppose $Y$ is logistic in $X$ conditioned on class with $\mathbb{P}(Y = 1 \mid X, A) = \sigma(\beta_A^T X)$, where $\sigma(t) = 1/(1 + e^{-t})$ and $\beta_0 = (1, -1)$, $\beta_1 = (1.25, -1)$ so that $X_2$ is predictive of creditworthiness in both classes but $X_1$ is slightly more predictive in class $A = 1$ than in $A = 0$. Suppose the training data comes from historically approved loans where loans were approved based on $X$ in such a way that $\mathbb{P}(Z = 1 \mid X) = \sigma(\beta_a^T X)$.

---

[1]The setting is motivated by systematic associations found in studies of the credit scores suggesting disadvantages for younger applicants and recent immigrants due to policies incorporating number of accounts (Board of Governors of the Federal Reserve System, 1997; Rice & Swesnik, 2014).

In Figs. 2a–2b we consider deriving a fair classifier for loan approval from the class-blind Bayes optimal score $\hat{R} = \mathbb{P}(Y = 1 \mid X, T = 1) = \mathbb{P}(Y = 1 \mid X, Z = 1)$, which is the same in training and target populations by construction and which we assume is given (*e.g.*, it can be identified from the training data regardless of any covariate shift between $Z = 1$ and $T = 1$; see Sec. 4.2). We simulate $n = 100000$ data points and censor the outcome for those with $Z_i = 0$. First we consider deriving an adjusted predictor from the Bayes-optimal classifier $\hat{Y} = \mathbb{I}[\hat{R} \geq 0.5]$ by naïvely applying the method of Hardt et al. (2016). Fig. 2a shows the space of achievable FPR-TPR in the training (censored) and target (full) populations along with the optimal equalized odds and equal opportunity rates corresponding to the symmetric loss $\ell(y, y') = \mathbb{I}[y \neq y']$. As can be seen, there is a significant discrepancy between the regions in the censored vs. full population. Next, we consider deriving optimal equal opportunity policies from the score $\hat{R}$. Fig. 2b shows the range of optimal policies, which is given by class-based thresholds, as we range the exchange rate between type-I and -II errors (false positives vs. false negatives). We also show the result from using a reweighting approach that we discuss in Sec. 5. We note that a naïve application of fairness adjustment provides *insufficient* compensation for the unfairness toward the underprivileged class $A = 1$: for every threshold on class $A = 0$, the corresponding threshold on class $A = 1$ is always *higher* for the policy derived in the naïve manner compared to that derived either using the full data or using our reweighting approach, such that the spuriously fair policy is systematically and uniformly harsher than necessary on the disadvantaged class.

### 4.1. Disparate Benefit of the Doubt

We now formalize the phenomenon illustrated in the example as *residual unfairness* and study why and when it arises in terms of the biases in training data due to existing prejudiced policies. For concreteness, we focus on the equality of opportunity criterion. Many of our results can be easily extended to other observational fairness criteria. To quantify the extent to which the criterion is satisfied or violated, and in which direction, we define the inequity of opportunity.

**Definition 2** (Inequity of Opportunity)**.** The inequity of opportunity between classes $A = a$ and $A = b$ wrt to event $E$ under policy $\hat{Y}$ is defined as

$$\epsilon_{a,b}^E = \mathbb{P}(\hat{Y} = 1 \mid {}^{E,A=a,}_{Y=1}) - \mathbb{P}(\hat{Y} = 1 \mid {}^{E,A=b,}_{Y=1})$$

Positive values in the target population, $\epsilon_{a,b}^{T=1} > 0$, indicate unfairness against group $b$. Zero inequity between all groups corresponds to equality of opportunity. A policy that is adjusted to be equal-opportunity or equalized-odds fair on the training data has $\epsilon_{a,b}^{Z=1} = 0$. Thus, any nonzero value of $\epsilon_{a,b}^{T=1}$ for such a policy constitutes a residual unfairness corresponding to the additional unadjusted-for inequity introduced by going from the $Z = 1$ to the $T = 1$ population.

Intuitively, if censoring induces a spuriously higher or lower overall distribution of scores than in the true population, we might learn a higher or lower threshold from the training data. If the true distribution will have more people with comparatively lower scores, the rate of false negatives will increase in the true population. This is to be expected if the censoring decision $Z = 1$ has itself an associated risk or cost, such as giving a loan. Differences in the extent and effects of this censoring between groups, which is what we will define as *disparate benefit of the doubt*, can then give rise to non-zero residual inequity. This may occur if the censoring mechanism subjects the disadvantaged group to harsher screening than the advantaged group, so that disadvantaged screened-in individuals have higher probabilities of being positive (e.g., innocent or creditworthy) given the observables $X, A$.

We next derive three sufficient conditions for residual unfairness. We explain how these can be interpreted in terms of the logging policy $Z = 1$ bestowing disparate benefit of the doubt with respect to the positive outcome $Y = 1$ on the different groups. This disparate benefit of the doubt would be directly reflected in the distribution of risk scores in the training and target populations, which we will show necessarily leads to residual unfairness that disadvantages the same group that received comparatively lesser benefit of the doubt (or comparatively heightened suspicion), thus perpetuating historical prejudices.

We first state a simple rephrasing of the residual inequity of opportunity left by a fairness adjustment. Recall that $F_a^{Z=1}(\theta), F_a^{T=1}(\theta)$ correspond to the false negative rate (FNR) when thresholding at $\theta$ on the training (censored) and test populations, respectively. Let $\Delta_a(\theta) = F_a^{Z=1}(\theta) - F_a^{T=1}(\theta)$ denote the difference between true positive rates in the test population and training (censored) population. Recall that $\hat{Y}$ is an optimal derived equal opportunity classifier based on the training data if $\hat{Y} = \mathbb{I}[\hat{R} > \theta_A]$ and $F_a^{Z=1}(\theta_a) = F_b^{Z=1}(\theta_b)$ for every two groups $a, b$.[2] For short we refer to such a classifier as a derived equal oppor-tunity classifier.

---

[2]Specifically, that the set of derived equal opportunity classi-

**Proposition 1.** *Let $\hat{Y} = \mathbb{I}[\hat{R} > \theta_A]$ be a derived equal opportunity classifier. Then $\epsilon_{a,b}^{T=1} = \Delta_a(\theta_a) - \Delta_b(\theta_b)$.*

Next, we define first-order stochastic dominance, which we use to express our first characterization of disparate benefit of the doubt.

**Definition 3** (First-order stochastic dominance). *Let $F$, $G$ be two CDFs. We write $F \preceq G$ whenever $F(\theta) \geq G(\theta) \; \forall \theta$.*

CDFs describe the distribution of a population of real values. The stochastic dominance $F \preceq G$ means that the population described by $F$ has overall smaller values than the population described by $G$. Specifically, $F \preceq G$ is equivalent to saying that for each unit in the population described by $F$ we can find a nonnegative number such that, when added to each unit, the whole population looks like that described by $G$ (Mas-Colell et al., 1997). That is, each unit from $F$ can be uniquely paired with a unit from $G$ such that the former has a smaller or equal value than the latter (allowing fractional or infinitesimal "units"). Alternatively, $F \preceq G$ holds if and only if the average of every increasing function in the $F$ population is smaller than or equal to the corresponding average in the $G$ population (Fishburn, 1980). This says that any rational actor with an increasing utility function would gain utility in choosing $G$ over $F$ and lose utility in choosing $F$ over $G$ (or get the same utility).

**Proposition 2** (Strong disparate benefit of the doubt). *Suppose that*

$$F_a^{Z=1} \preceq F_a^{T=1} \quad \text{and} \quad F_b^{Z=1} \succeq F_b^{T=1} \qquad (2)$$

*while not both are equalities, i.e., either $F_a^{Z=1} \neq F_a^{T=1}$ or $F_b^{Z=1} \neq F_b^{T=1}$ (or both). Then every derived equal opportunity classifier has nonnegative inequity of opportunity for group $b$ relative to group $a$ ($\epsilon_{a,b}^{T=1} \geq 0$) and at least one derived equal opportunity classifier will have a strictly positive inequity of opportunity disadvantaging group $b$ relative to group $a$ ($\epsilon_{a,b}^{T=1} > 0$).*

The condition in eq. (2) requires that the distribution of scores among positive group-$a$ members is overall smaller in the training data than in the target population, while the opposite is true for group $b$. Recall that scores represent the probability of having the positive, favorable label (more on this in Sec. 4.2). Thus, the condition says that positive group-$a$ members received *more* benefit of the doubt when

---

fiers that are optimal with respect to some trade off between type-I and -II errors is exactly equal to the set of all such thresholding classifiers requires only that we assume that, in each group $A = a$, $\hat{R}$ is not worse than random guessing and that the ROC is convex. Neither is without loss of generality as seen by only improving $\hat{R}$ by conditionally (on $A$) negating $\hat{R}$ and/or randomizing its value in nonconvex intervals between the endpoints.

being screened-in into the training data than positive group-$b$ members. Prop. 2 shows that this will necessarily lead to group-$b$ being further disadvantaged in the future even after correcting for equality of opportunity.

In the context of loan application, where we can think about the score as a credit score, eq. (2) means that the logging policy (*i.e.*, historical loan approval practice) effectively dug deeper into the pile of creditworthy group-$a$ applicants than for group-$b$ applicants, giving the former more benefit of the doubt as to their creditworthiness based on their credit scores than it gave the latter. Seen via the equivalent utility-based interpretation of stochastic dominance, given any increasing utility function, if eq. (2) holds then the logging policy is losing utility on group-$a$ via lax screening while gaining utility on group-$b$ by being less lax.

The complement (or, negative) of the score can be thought of as a *risk score*: the probability of the unfavorable label. Eq. (2) can equivalently be written as the opposite ordering on risk scores rather than positivity scores. Thus, in either a judicial bail- or sentence-setting context or in a predictive policing context, eq. (2) means that the logging policy (*i.e.*, historical criminal justice or policing practice) was harsher on group $b$ than on group $a$, screening-in *lower* risk scores for innocent group-$b$ members compared to the group-$b$ population while screening-in only *higher* risk scores for group-$a$ members, giving them more benefit of the doubt as to their innocence based on observables.

If the CDFs are nowhere equal except for at 0 and 1 (where they are always equal) then a strict version of Prop. 2 shows that *every* derived equal opportunity classifier will be unfair.

**Proposition 3** (Strong disparate benefit of the doubt, strict). *Suppose eq. (2) holds and that $F_a^{Z=1}(\theta) \neq F_a^{T=1}(\theta)$, $F_b^{Z=1}(\theta) \neq F_b^{T=1}(\theta)$ for all $\theta \in (0,1)$. Then every nontrivial derived equal opportunity classifier will have strictly positive inequity of opportunity disadvantaging group $b$ relative to group $a$.*

A nontrivial classifier is any classifier that is neither the constant $\hat{Y} = 0$ nor the constant $\hat{Y} = 1$.

The conditions in Props. 2 and 3 are easy to interpret via stochastic dominance but may be too strong to hold in practice. In particular, suppose the decision $Z = 1$ itself has a benefit or risk related to whether $Y = 1$, as in the case of giving a loan (benefit of earning the full interest over loan term compared to risk of a default) or a police stop (benefit of curtailing crime compared to costs, including societal, of aggressive policing). Then, *if* the decision $Z = 1$ is exercised rationally, then we would expect that the distribution of scores is either overall higher (*e.g.*, for loans) or overall lower (*e.g.*, for police stops) in the training population regardless of group, *i.e.*, both $F_a^{Z=1} \succeq F_a^{T=1}$ and $F_b^{Z=1} \succeq F_b^{T=1}$ or both $F_a^{Z=1} \preceq F_a^{T=1}$ and $F_b^{Z=1} \preceq F_b^{T=1}$. (Al-

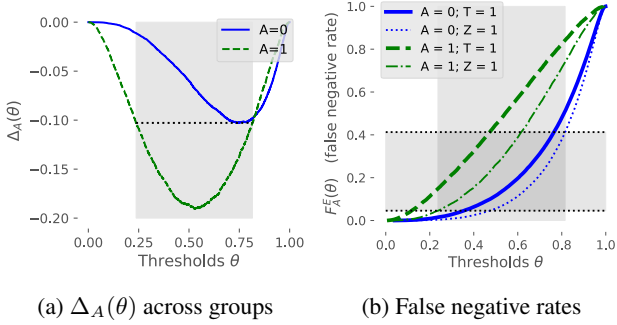

(a) $\Delta_A(\theta)$ across groups  (b) False negative rates

Figure 3: Residual unfairness against due to weak disparate benefit of the doubt in the loan application example.

though, prejudice in $Z=1$ can be so overt and/or irrational for this not to hold.) If this is the case, then the conditions of Props. 2 and 3 cannot hold and we must relax them.

The next result shows that even if the stochastic dominance holds in the same direction for both groups, if the *magnitude* of the dominance is overall larger in one group compared to the other for a large swath of thresholds then most derived equal opportunity classifiers will actually be unfair and disadvantage the historically disadvantaged.

**Proposition 4** (Weak disparate benefit of the doubt, strict). *Let $\underline{\theta}, \overline{\theta}$ be such that*

$$\Delta_a(\theta) > \Delta_b(\theta') \quad \forall \theta, \theta' \in (\underline{\theta}, \overline{\theta}). \quad (3)$$

*Let $\hat{Y} = \mathbb{I}[\hat{R} > \theta_A]$ be a derived equal opportunity classifier. If $\theta_a, \theta_b \in (\underline{\theta}, \overline{\theta})$, then $\hat{Y}$ induces a strictly positive inequality of opportunity disadvantaging group b relative to group a.*

Note that, since $\Delta_A(0) = \Delta_A(1) = 0$, we have that eq. (3) holds for $(\underline{\theta}, \overline{\theta}) = (0, 1)$ if and only if the conditions of Prop. 3 hold. Therefore, for general $(\underline{\theta}, \overline{\theta})$ the former can be understood as a relaxation of the latter.

We can illustrate the conditions of Prop. 4 in the synthetic loan application example from the beginning of this section. In Fig. 3b, we plot the two $\Delta_A$ functions and shade a large interval where eq. (3) holds. In Fig. 3a, we plot the CDFs $F_A^E$ and further shade the corresponding regions of false negative rates for which both $\theta_0$ and $\theta_1$ lie the previously shaded interval. Taking complements, this shows that *any* derived equal opportunity classifier adjusted to have equal true positive rates of 0.58–0.95 on the training data will disadvantage the underprivileged class despite one's attempt to adjust against this situation.

We can slightly relax the condition in eq. (3) whenever dealing with groups that have disparate endowments of scores, as in the example above where the scores of group 0 are overall larger than those of group 1 in terms of stochastic dominance.

**Proposition 5** (Weak disparate benefit of the doubt on disparately endowed groups). *Let $\underline{\theta}, \overline{\theta}$ be such that*

$$\Delta_a(\theta) > \Delta_b(\theta') \quad \forall \theta, \theta' \in (\underline{\theta}, \overline{\theta}) : \theta \geq \theta'. \quad (4)$$

*Suppose $F_a^{Z=1} \succeq F_b^{Z=1}$. Let $\hat{Y} = \mathbb{I}[\hat{R} > \theta_A]$ be a derived equal opportunity classifier. If $\theta_a, \theta_b \in (\underline{\theta}, \overline{\theta})$, then $\hat{Y}$ induces a strictly positive inequality of opportunity disadvantaging group b relative to group a.*

In the supplemental Sec. B we also include an illustration of weak disparate benefit of the doubt in a real dataset of credit card applications and payment defaults. In addition to making more concrete our crediting example, this also serves to illustrate the weaker condition in eq. (4).

All of our results in this section can be equivalently stated for true negative rates instead of true positive rates, in which case the corresponding conditions such as that in eq. (2) can instead be interpreted as *disparate suspicion*, i.e., the disparate scrutiny of truly criminal or credit-unworthy individuals. So, whereas our notion of disparate benefit of the doubt corresponds to the phenomenon of "driving while black" (Lamberth, 1998), our notion of disparate suspicion would correspond to the phenomenon of "criming while white" (Goldfarb, 2014). If *either* disparate benefit of the doubt *or* disparate suspicion is present, a derived equalized odds classifier will in fact *violate* equalized odds in a way that disadvantages the same group that was disadvantaged by the disparate benefit of the doubt or suspicion.

### 4.2. Interpretation of Scores Under MAR

In the above we interpreted the conditions in our results as disparities in the distributions of positivity or risk scores among different groups in the training and target populations. Specifically, we interpreted these scores as corresponding to the probabilities of having a positive or negative label given observables. In full generality, however, this probability might actually be different in the training and target populations, i.e., $\mathbb{P}(Y=1 \mid X, A, Z=1) \neq \mathbb{P}(Y=1 \mid X, A, T=1)$. Since naturally only the training data is available at training we can consider the training-population Bayes score $\hat{R} = \mathbb{P}(Y=1 \mid X, A, Z=1)$ and interpret disparities as disparities in benefit of the doubt of positivity given this score. This is consistent with our interpretation above.

However, whenever censoring $Z=1$ is itself based on observables, the data will be *missing conditionally at random* (MAR) and these probabilities will actually be the same so that we can interpret the scores as simply the probability of positivity given observables generally.

**Assumption 1.** (MAR) $Z \perp\!\!\!\perp Y \mid X, A$ and $T \perp\!\!\!\perp Y \mid X, A$.

Under MAR, it is immediate that the Bayes score satisfies $\hat{R} = \mathbb{P}(Y=1 \mid X, A) = \mathbb{P}(Y=1 \mid X, A, Z=1) =$

$\mathbb{P}(Y = 1 \mid X, A, T = 1)$, which can be consistently estimated from training data. In fact, under MAR, the optimal (unrestricted) decision function in $X, A$ minimizing the average over $Z = 1$ of any loss in $Y$ is the same as that minimizing the average loss over $T = 1$.

MAR requires that the missingness is unrelated to outcome after controlling for the observables. This assumption is clearly satisfied in the common case when $T = 1$ is constant and only $X, A$ (or just $X$) were taken into consideration for a randomized inclusion policy $Z$. In the examples laid out in Sec. 1, this is an appropriate assumption because the censoring mechanism does not observe outcomes $Y$ a priori, only observable characteristics (including the protected attribute). However, violations of MAR may occur, for example in the loan case if applicants may choose an outside option, self-censoring the observation of a default while the availability of these options is related to creditworthiness.

## 5. Fairness Assessment and Adjustment with Biased Data via Sample Reweighting

In rare situations, we may have additional information about the target population such as an unlabeled dataset. We next show how we can use such data to evaluate accuracy metrics on the target population, as long as data is MAR. This will allow us to assess the residual unfairness of fairness adjusted classifiers, as we will do in our study of SQF, and to correctly adjust for fairness, as we have done in Fig. 2b.

Let $p(x, a) = \frac{\mathbb{P}(T=1|X=x,A=a)}{\mathbb{P}(Z=1|X=x,A=a)}$ be the propensity score ratio between the target and training populations. This ratio is a standard way to adjust for systematic covariate shift for evaluating averages in the target (Horvitz & Thompson, 1952; Bottou et al., 2012). We next state how a weighting score that's equal to it up to proportionality in $a$ can be used for evaluating true positive rates.

**Proposition 6.** *Suppose $\tilde{p}(x, a) = r(a)p(x, a)$ for some $r(a)$ and $\hat{Y} \perp (Y, Z, T) \mid X, A$. Then, under Assumption 1, $\mathbb{P}(\hat{Y} = 1 \mid Y = y, A = a, T = 1)$ is equal to*

$$\frac{\mathbb{E}[\mathbb{I}[\hat{Y}=1,Y=y,A=a]\tilde{p}(X,A)|Z=1]}{\sum_{\hat{y}\in\{0,1\}} \mathbb{E}[\mathbb{I}[\hat{Y}=\hat{y},Y=y,Z=1,A=a]\tilde{p}(X,A)|Z=1]} \quad (5)$$

Given $\tilde{p}(x, a)$, the quantity in eq. (5) involves only the distribution of the training data. In practice, the expectations in it can be estimated using empirical averages over the training data. Therefore, *if* we can compute an appropriate weighting function $\tilde{p}(x, a)$, Prop. 6 provides a remedy to the problem of biased data: we may simply replace the condition in Def. 1, which involves an unknown distribution $X, A, Y \mid T = 1$, with the condition that eq. (5) is constant over $a$ (for $y \in \{0, 1\}$ or for $y = 1$). In particular, to apply the equality of opportunity adjustment on the target population, we need only compute the TPRs and/or FPRs of any blackbox predictor using the adjustment of eq. (5) and proceed with the adjustment as usual.

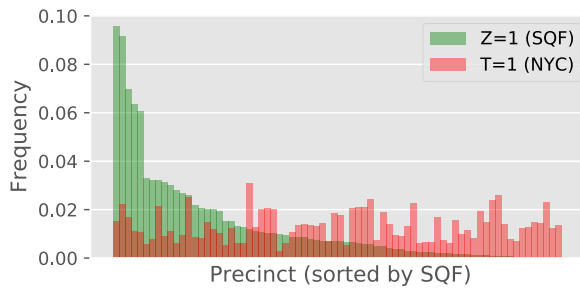

Figure 4: Precincts by SQF stops vs. by NYC population

Next we address when can we find an appropriate reweighting score $\tilde{p}(x, a)$. We consider two cases. If $Z = 1$ is a subpopulation of $T = 1$ and our data consists of iid draws from the target population population but where, naturally, only the $Z = 1$ units are labeled, then may simply let $\tilde{p}(x, a)$ be the reciprocal of the conditional probability of being labeled, which may be estimated using a probabilistic classification algorithm such as logistic regression. This case applies, for example, in the loan approval policy example if the data includes the full loan applications, whether they were approved, and whether the approved loans defaulted or were paid back. If, however, our data consists only of the labeled examples, as in the case of arrest and SQF data, which contain only the arrests or stops made and, naturally, never any information on those *not* made, then this case does not apply. But, if we have an unlabeled dataset from the target population, then we can separately estimate the distribution of $X, A$ in the training and target distributions. Then we may let $\tilde{p}(x, a)$ be either the ratio of densities of $X, A$ in $T = 1$ and $Z = 1$ or the ratio of densities of $X$ in $T = 1, A = a$ and $Z = 1, A = a$.

In supplemental Sec. C we provide additional results characterizing residual unfairness under MAR in terms of $\tilde{p}(x, a)$.

## 6. Case Study: Stop, Question, and Frisk

We next study the Stop, Question, and Frisk dataset to illustrate how residual unfairness may occur with real data. We consider learning a predictor of criminal possession of a weapon from this dataset using logistic regression and then adjusting the policy to be fair on the training population. The trained policy will be applied to the target population of New York City at large, where it will be shown to in fact be unfair. This residual unfairness can be explained by disparities between the training population (SQF stops) and the target population (NYC), which is evident from the divergent geographic distributions of these two as seen in Fig. 4. (Further details about SQF and the dataset are provided in the supplemental Sec. D.)

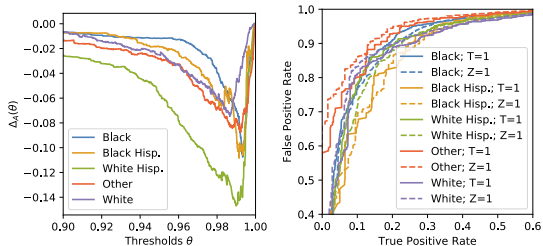

Figure 5: Disparate benefit of the doubt in SQF

Table 2: Residual unfairness in SQF predictive targeting

|  | Equal Opportunity | | | | Equalized Odds | | | |
|  | $Z=1$ | | $T=1$ | | $Z=1$ | | $T=1$ | |
|  | FNR | FPR | FNR | FPR | FNR | FPR | FNR | FPR |
| --- | --- | --- | --- | --- | --- | --- | --- | --- |
| Black | 0.11 | 0.19 | 0.14 | 0.13 | 0.14 | 0.20 | 0.17 | 0.15 |
| Black Hisp | 0.11 | 0.26 | 0.15 | 0.22 | 0.14 | 0.20 | 0.19 | 0.15 |
| White Hisp | 0.11 | 0.22 | **0.20** | 0.11 | 0.14 | 0.20 | **0.22** | 0.12 |
| Other | 0.11 | 0.13 | 0.16 | 0.12 | 0.14 | 0.20 | 0.17 | 0.18 |
| White | 0.11 | 0.19 | **0.11** | 0.20 | 0.14 | 0.20 | **0.15** | 0.22 |

Goel et al. (2016) considered learning predictors from this data as a way to assess the implied decision thresholds used for determining pedestrian stops related to the criteria of "reasonable suspicion." While the authors suggest that a logistic regression predictor of criminal possession of a weapon could be used as a secondary filter applied on those individuals targeted by officers, we instead consider training such a predictive model to guide whom to target for a stop and search and adjusting this targeting policy for fairness.

We consider race to be the protected class $A$, with values Black, Black Hispanic, White, White Hispanic, and Other.[3] Arrest data or SQF data only contain the arrests or stops made and, naturally, never any information on those *not* made. Since biases in the demographics of stop data arise from disproportionate policing by location and potential racial biases, we define the target population with respect to demographic data about NYC precincts. Letting $X_1$ denote the precinct-encoding portion of the covariates and $X_2$ denote all other covariates (which include 30 indicators of reasons for suspicion, sex, and location or stop), we set $\mathbb{P}(X_1, A \mid T=1)$ to be the same distribution as that of the population of NYC. We further assume that, once we condition on the main sources of bias, $(X_1, A)$, the covariates $X_2$ are not disproportionate between the training and target distribution: we set $\mathbb{P}(X_2 \mid A, X_1, T=1) = \mathbb{P}(X_2 \mid A, X_1, Z=1)$. That is, the main sources of systematic bias with respect to censored data are contained in $X_1$ and $A$, while ancillary covariates $X_2$ are not proxies for discrimination. We then have that $\tilde{p}(X, A) = \frac{\mathbb{P}(A, X_1 \mid T=1)}{\mathbb{P}(A, X_1 \mid Z=1)}$ satisfies the conditions of Prop. 6 so we need only estimate $\mathbb{P}(A, X_1 \mid Z=1)$, $\mathbb{P}(A, X_1 \mid T=1)$. We estimate the former from the SQF data and the latter from the 2010 American Community Survey data by matching census blocks to precincts and using Laplace smoothing. We clip the ratio weights at 10.

Using the SQF data as is, we fit a logistic regression $\hat{R}$ to predict the probability of *innocence*, that a search would not recover a weapon from those suspected of criminal possession one based on the covariates $X$. Fig. 5 shows the FNR discrepancies between training and target and the ROCs in each. We consider deriving both an equal opportunity classifier and an equalized odds classifier for a stop and search based on the SQF data to minimize false negatives and false positives at an exchange rate of 25:1.

In Table 2, we report the estimated true positive and false positive rates achieved by these classifiers both in the training and in the target population. The latter is computed using Prop. 6. FNRs quantify the percent of innocents wrongly targeted and FPRs quantify the percent of criminals undetected. By construction, the adjusted-for rates are equal in the training population ($Z=1$). However, in practice, residual unfairness remains in the target population even after adjustment, and both of these supposedly fair classifiers will systematically disadvantage the same groups that were previously disparately targeted. For the equal-opportunity-adjusted classifier, whereas only 11% of white-non-Hispanic innocents are wrongly targeted, up to 20% of white-Hispanic, 16% of other, and 14–15% of black innocents are wrongly targeted and harassed. Similar disparities exist for the equalized-odds-adjusted classifier. The equalized odds policy, having been subject to additional constraints that require more randomization and hence less dependence on observables, induces smaller but still significant disparities and Hispanics remain particularly disproportionately burdened.

In all cases, after fairness adjustment, non-white individuals were still unfairly disadvantaged in practice relative to white individuals, thus perpetuating the same biases that SQF is notorious for under the guise of a policy adjusted to be fair.

## 7. Conclusion

Our work characterizes the problem of residual unfairness, which arises when policies learned from biased datasets are adjusted for fairness but remain unfair in practice. We study a general setting where the dataset is generated under a prejudiced historical policy, which captures the structure of many problem settings where fairness has been considered. We prove that the same prejudices will be reflected in the supposedly-fairness-adjusted policy.

---

[3] Due to the relative size of the Asian/Pacific Islander and Native American classes included in the original SQF dataset, we combine them with the Other (U) class.

## A. Omitted Proofs

*Proof of Prop. 1.* Let $1 - \rho = F_a^{Z=1}(\theta_a) = F_b^{Z=1}(\theta_b)$, which are equal by assumption. We then have that

$$\begin{aligned}
\epsilon_{a,b} &= (1 - F_a^{T=1}(\theta_a)) - (1 - F_b^{T=1}(\theta_b)) \\
&= F_b^{T=1}(\theta_b) - F_a^{T=1}(\theta_a) \\
&= ((1-\rho) - F_a^{T=1}(\theta_a)) - ((1-\rho) - F_b^{T=1}(\theta_b)) \\
&= (F_a^{Z=1}(\theta_a) - F_a^{T=1}(\theta_a)) - (F_b^{Z=1}(\theta_b) - F_b^{T=1}(\theta_b)) \\
&= \Delta_a(\theta_a) - \Delta_b(\theta_b).
\end{aligned}$$
□

*Proof of Prop. 2.* Let $\hat{Y} = \mathbb{I}[\hat{R} > \theta_A]$ be any derived equal opportunity classifier. By Prop. 1, $\epsilon_{a,b} = \Delta_a(\theta_a) - \Delta_b(\theta_b)$ and by assumption $\Delta_a(\theta_a) \geq 0$ while $\Delta_b(\theta_b) \leq 0$.

By assumption, at least one of $F_a^{Z=1} \neq F_a^{T=1}$ and $F_b^{Z=1} \neq F_b^{T=1}$ holds. Suppose that $F_a^{Z=1} \neq F_a^{T=1}$. Then there is $\theta_a$ such that $F_a^{Z=1}(\theta_a) \neq F_a^{T=1}(\theta_a)$. Then $\Delta_a(\theta_a) > 0$. Letting $\theta_b = (F_b^{Z=1})^{-1}(F_a^{Z=1}(\theta_a))$, we get that $\hat{Y} = \mathbb{I}[\hat{R} > \theta_A]$ is a derived equal opportunity classifier with $\epsilon_{a,b} > 0$. If instead $F_b^{Z=1} \neq F_b^{T=1}$ then we'd have $\theta_b$ with $\Delta_b(\theta_b) < 0$ and we'd let $\theta_a = (F_a^{Z=1})^{-1}(F_b^{Z=1}(\theta_b))$.
□

*Proof of Prop. 3.* Let $\hat{Y} = \mathbb{I}[\hat{R} > \theta_A]$ be any nontrivial derived equal opportunity classifier. By Prop. 1, $\epsilon_{a,b} = \Delta_a(\theta_a) - \Delta_b(\theta_b)$ and by assumption $\Delta_a(\theta_a) > 0$ and $\Delta_b(\theta_b) < 0$.
□

*Proof of Prop. 4.* Self-evident from Prop. 1. □

*Proof of Prop. 5.* Self-evident from Prop. 1 after noting that $F_a^{Z=1} \succeq F_b^{Z=1}$ necessarily implies that $\theta_a \geq \theta_b$. □

*Proof of Prop. 6.* We have that

$$\begin{aligned}
\mathbb{P}(\hat{Y}=\hat{y}, Y=y, A=a \mid T=1) &= \mathbb{E}[\mathbb{P}(\hat{Y}=\hat{y}, Y=y, T=1 \mid X, A)\mathbb{I}[A=a]]/\mathbb{P}(T=1) \\
&= \mathbb{E}[\mathbb{P}(\hat{Y}=\hat{y}, Y=y \mid X, A, T=1)\mathbb{P}(T=1 \mid X, A)\mathbb{I}[A=a]]/\mathbb{P}(T=1) \\
&= \mathbb{E}[\mathbb{P}(\hat{Y}=\hat{y} \mid X, A, T=1)\mathbb{P}(Y=y \mid X, A, T=1)\mathbb{P}(T=1 \mid X, A)\mathbb{I}[A=a]]/\mathbb{P}(T=1) \\
&= \mathbb{E}[\mathbb{P}(\hat{Y}=\hat{y} \mid X, A)\mathbb{P}(Y=y \mid X, A)\mathbb{P}(T=1 \mid X, A)\mathbb{I}[A=a]]/\mathbb{P}(T=1) \\
&= \mathbb{E}[\mathbb{P}(\hat{Y}=\hat{y} \mid X, A, Z=1)\mathbb{P}(Y=y \mid X, A, Z=1)\mathbb{P}(T=1 \mid X, A)\mathbb{I}[A=a]]/\mathbb{P}(T=1) \\
&= \mathbb{E}[\mathbb{P}(\hat{Y}=\hat{y}, Y=y \mid X, A, Z=1)\mathbb{P}(T=1 \mid X, A)\mathbb{I}[A=a]]/\mathbb{P}(T=1) \\
&= \mathbb{E}[\mathbb{P}(\hat{Y}=\hat{y}, Y=y, Z=1 \mid X, A)\tfrac{\mathbb{P}(T=1|X,A)}{\mathbb{P}(Z=1|X,A)}\mathbb{I}[A=a]]/\mathbb{P}(T=1) \\
&= \mathbb{E}[\mathbb{I}[\hat{Y}=\hat{y}, Y=y, Z=1, A=a]p(X,A)]/\mathbb{P}(T=1).
\end{aligned}$$

The rest follows by Bayes law. □

## B. Weak Disparate Benefit of the Doubt in Credit Card Data

We consider the data from Greene (1992), which contains individual-level data on credit card acceptance, default on payments or not (if accepted), information about individual income, derogatory reports on accounts, and other features of creditworthiness such as self-employment indicators. The dataset also includes age, which has been a concern regarding the fairness of credit scoring models (Board of Governors of the Federal Reserve System, 1997). We construct a protected class by defining $A = \mathbb{I}[X_{age} < F_{X_{age}}^{-1}(0.5)]$ as the indicator for being below the median age (31.67). To illustrate how the direction of disparities can change depending on the $Z=1$ policy we consider two scenarios. In both, we consider the target population $T=1$ to actually consist of all accepted applicants (rather than all applicants). First, we consider $Z = \mathbb{I}[T=1, X_i > F_{X_{inc}}^{-1}(.1)]$ where $X_{inc}$ is income so that we *further* censor the lowest-income individuals from the available data. Second, we also consider $Z = \mathbb{I}[T=1, X_i > F_{X_{inc\_per}}^{-1}(.1)]$ where $X_{inc\_per}$ is the income per dependent. In $T=1$, income is somewhat correlated with membership in the protected class (being young), with a correlation coefficient $\rho = -0.32$. However, income per dependent is very weakly correlated with being young, with a correlation coefficient

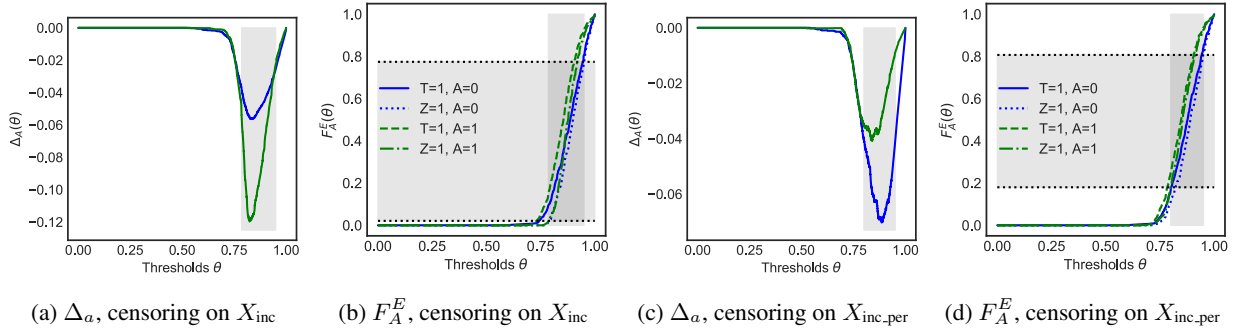

(a) $\Delta_a$, censoring on $X_{\text{inc}}$  (b) $F_A^E$, censoring on $X_{\text{inc}}$  (c) $\Delta_a$, censoring on $X_{\text{inc\_per}}$  (d) $F_A^E$, censoring on $X_{\text{inc\_per}}$

Figure 6: Illustration of score disparities between censored and full training data. The dataset is from credit card applications, where we additionally censor the population of accepted cardholders by income or income per dependent.

$\rho = 0.03$. This is intuitive as the correlation of greater income by age is canceled out by the correlation of greater household size with age.

In Fig. 6, we plot the FNRs $F_A^{T=1}$, $F_A^{Z=1}$ in training and in target and the discrepancies $\Delta_A$ between them for both censoring cases. First we study the case of censoring on $X_{\text{inc}}$. By inspecting the FNRs in Fig. 6b, we see that $F_0^{Z=1} \succeq F_1^{Z=1}$. Therefore, we can apply Prop. 5. In Fig. 6a we shade a region of thresholds that satisfies eq. (4). In particular, because of the relaxation for disparately endowed groups, we can extend the region farther right than would be possible under eq. (3). In Fig. 6b we shade the corresponding ranges of FNRs that, per Prop. 5, would lead to spuriously-fairness-adjusted classifiers that actually induce an inequity of opportunity disadvantages the younger group $A = 1$.

Next we study the case of censoring on $X_{\text{inc\_per}}$. We first note that the disparate benefit of the doubt induced is now going in the *opposite* direction (Fig. 6c) – so the spuriously-fairness-adjusted classifiers will disadvantage the older group rather than the younger group. Although we have the same ordering of FNRs as before, $F_0^{Z=1} \succeq F_1^{Z=1}$ (Fig. 6d), the ordering of $\Delta_A$'s is opposite and therefore eq. (4) does not offer a relaxation over eq. (3). We therefore apply the standard weak disparate benefit of the doubt. In Fig. 6c we shade a region of thresholds that satisfies eq. (3). In Fig. 6b we shade the corresponding ranges of FNRs, per Prop. 4, would lead to spuriously-fairness-adjusted classifiers that actually induce an inequity of opportunity disadvantages the older group $A = 0$.

## C. Residual Unfairness Under MAR

In this section we study several implications for residual unfairness under the MAR assumption. First we show that in rare cases prejudice, if applied purely and directly on protected attribute alone, can actually be perfectly corrected for based on training-data-based fairness adjustment. Second we study how disparities in importance weights can be used to characterize the presence residual unfairness.

### C.1. No Residual Unfairness if Inclusion Depends Only on Protected Attributes

We next show that if the censoring mechanism depends *only on the protected attribute A* that is to be adjusted for fairness, then in fact there will be no residual unfairness and the true positive rates remain the same in the target population as in the training population.

**Proposition 7.** *Suppose* $\mathbb{P}(Z = 1 \mid X, A) = \mathbb{P}(Z = 1 \mid A)$ *and* $\mathbb{P}(T = 1 \mid X, A) = \mathbb{P}(T = 1 \mid A)$. *Then, under Assumption 1, inequity of opportunity on training is the same as inequity of opportunity on target, i.e.,* $\epsilon_{a,b}^{Z=1} = \epsilon_{a,b}^{T=1}$.

Thus, residual unfairness occurs only when biased inclusion is heterogeneous based on covariates $X$, *e.g.*, as in the case of SQF where the application of stops differs based on both precinct and race. This is typical of the application areas where fairness is of concern: censoring is usually disparate in large part via proxies for protected attributes.

### C.2. Characterizing Residual Unfairness in Terms of Propensity Ratios Disparities

Under Assumption 1, we can characterize residual unfairness in terms of the reweighting estimates.

**Proposition 8.** $\Pr[\hat{Y} = 1 \mid {}^{Y=1}_{A=a,T=1}] > \Pr[\hat{Y} = 1 \mid {}^{Y=1}_{A=a,Z=1}]$ if and only if

$$\mathbb{E}\left[p(X,A) \mid {}^{Z=1,A=a}_{Y=1,\hat{Y}=0}\right] < \mathbb{E}\left[p(X,A) \mid {}^{Z=1,A=a}_{Y=1,\hat{Y}=1}\right]$$

This says that the TPR will be in actuality higher in the target population if the average ratio weights in the group of true positives included in the dataset is greater than in the group of false negatives included. Intuitively, the predictor will be more accurate in the target population if, due to censoring, positive examples that the predictor will be correct on were less likely to appear in the training data.

We can use to characterize exactly when a classifier that satisfies equal opportunity on training will have residual unfairness on target.

**Corollary 9.** Suppose $\hat{Y}$ satisfies equal opportunity wrt $Z = 1$. Then $\epsilon_{a,b}^{T=1} > 0$ if and only if

$$\frac{\mathbb{E}\left[p(X,A) \mid {}^{Z=1,A=a}_{Y=1,\hat{Y}=0}\right]}{\mathbb{E}\left[p(X,A) \mid {}^{Z=1,A=b}_{Y=1,\hat{Y}=0}\right]} < \frac{\mathbb{E}\left[p(X,A) \mid {}^{Z=1,A=a}_{Y=1,\hat{Y}=1}\right]}{\mathbb{E}\left[p(X,A) \mid {}^{Z=1,A=b}_{Y=1,\hat{Y}=1}\right]}$$

This characterization follows from Prop. 8 and the fact that the true positive rates in training are the same under equality of opportunity.

### C.3. Proofs

*Proof of Prop. 7.* The result follows from applying Prop. 6 with $\Pr(Z = 1 \mid X, A) = \Pr(Z = 1 \mid A)$ and iterating the expectation over $X$:

$$\mathbb{P}(\hat{Y} = 1 \mid Y = y, A = a, T = 1)$$

$$= \frac{\mathbb{E}[\frac{1}{\Pr[Z=1|A]}\mathbb{E}[\mathbb{E}[\mathbb{I}\{\hat{Y}=1, Y=1, Z=1\} \mid X]]\mathbb{I}\{A=a\}]}{\sum_{\hat{y} \in \{0,1\}} \mathbb{E}[\frac{1}{\Pr[Z=1|A]}\mathbb{E}[\mathbb{E}[\mathbb{I}\{\hat{Y}=\hat{y}, Y=1, Z=1\} \mid X]]\mathbb{I}\{A=a\}]}$$

$$= \frac{\mathbb{E}[\frac{1}{\Pr[Z=1|A]}]\mathbb{E}[\mathbb{E}[\mathbb{E}[\mathbb{I}\{\hat{Y}=1, Y=1, Z=1\} \mid X]]\mathbb{I}\{A=a\}]}{\mathbb{E}[\frac{1}{\Pr[Z=1|A]}] \sum_{\hat{y} \in \{0,1\}} \mathbb{E}[\mathbb{E}[\mathbb{E}[\mathbb{I}\{\hat{Y}=\hat{y}, Y=1, Z=1\} \mid X]]\mathbb{I}\{A=a\}]}$$

$$= \mathbb{P}(\hat{Y} = 1 \mid Y = y, A = a, Z = 1) \qquad \square$$

*Proof of Prop. 8.*

$$\Delta_a^{TPR} = \frac{\text{TPR}^*}{\text{TPR}} = \frac{\Pr[\hat{Y}=1 \mid Y=1, A=a, T=1]}{\Pr[\hat{Y}=1 \mid Y=1, A=a, Z=1]}$$

$$= \frac{\frac{\Pr[\hat{Y}=1,y=1|A=a,T=1]/\Pr[Y=1|A=a,T=1]}{\sum_{\hat{y}\in\{0,1\}} \Pr[\hat{Y}=1,y=1|A=a,T=1]}}{\frac{\Pr[\hat{Y}=1,y=1|A=a,T=1]/\Pr[Y=1|A=a,Z=1]}{\sum_{\hat{y}\in\{0,1\}} \Pr[\hat{Y}=1,y=1|A=a,Z=1]}} = \frac{1 + \frac{\Pr[Y=1|A=a,T=1]}{\Pr[\hat{Y}=1,y=1|A=a,T=1]}}{1 + \frac{\Pr[Y=1|A=a,Z=1]}{\Pr[\hat{Y}=1,y=1|A=a,Z=1]}}$$

So

$$\Delta_a^{TPR} > 1 \iff \frac{\Pr[Y=1, \hat{Y}=0 \mid A=a, T=1]}{\Pr[Y=1, \hat{Y}=1 \mid A=a, T=1]} < \frac{\Pr[Y=1, \hat{Y}=0 \mid A=a, Z=1]}{\Pr[Y=1, \hat{Y}=1 \mid A=a, Z=1]}$$

We can apply Prop. 6:

$$\Delta_a^{TPR} > 1 \iff \frac{\Pr[Y=1, \hat{Y}=1 \mid A=a, Z=1]}{\mathbb{E}[\mathbb{I}[Y=1, \hat{Y}=1]p(X,A) \mid {}^{Z=1}_{A=a}]} < \frac{\Pr[Y=1, \hat{Y}=0 \mid A=a, Z=1]}{\mathbb{E}[\mathbb{I}[Y=1, \hat{Y}=0]p(X,A) \mid {}^{Z=1}_{A=a}]}$$

and the identification that $\mathbb{E}[\mathbb{I}\{Y=1, \hat{Y}=0\}p(X,A) \mid {}^{Z=1}_{A=a}] = \mathbb{E}\left[p(x,A) \mid {}^{Z=1,A=a}_{Y=1,\hat{Y}=0}\right]\Pr[Y=1, \hat{Y}=0 \mid {}^{Z=1}_{A=a}]$:

$$\Delta_a^{TPR} > 1 \iff \frac{1}{\mathbb{E}\left[p(X,A) \mid {}^{Z=1,A=a}_{Y=1,\hat{Y}=1}\right]} < \frac{1}{\mathbb{E}\left[p(X,A) \mid {}^{Z=1,A=a}_{Y=1,\hat{Y}=0}\right]} \qquad \square$$

# D. Information on Stop, Question and Frisk

Stop, Question, and Frisk is a program which allows police officers to stop citizens in public, question, and possibly search them, under reasonable suspicion of a crime but not enough probable cause for an arrest (Goel et al., 2017). Around 600,000 people were stopped in 2011, and around 90% of stops led to no evidence of a crime (Keefe, 2011). Each officer is required to file an individual report after a stop detailing individual characteristics (including physical attributes of the suspect and location) and reasons for the stop, leading to relatively rich context about each individual decision (NYCLU, 2017).

SQF was studied by statistical researchers and adjudicated in the court case Floyd v. City of New York for discrimination on the basis of race and national origin. The program has been controversial since the demographic makeup of stops in the data systematically misrepresents the population of NYC at large due to disparate patrol levels and implementation of SQF by NYPD precinct, which correlates with demographics, as well as the potential for racial biases at the individual level. These demographic imbalances have been studied and analyzed judicially, discussed alongside evidence of administrative and structural deviation in application of SQF practices (Goel et al., 2017; Gelman et al., 2007). Some of the covariates themselves may reflect proxy indicators for discrimination. In the SQF data, for example, recorded reasons for stop include whether the suspect was actually engaging in a crime, was a known criminal, or exhibited "furtive movement". The potential for some of these reasons to be proxies for discrimination was noted by Judge Scheindlin in the court case of Floyd, et al. v. City of New York.